\documentclass{article}

\usepackage{arxiv}
\usepackage{array}
\usepackage[utf8]{inputenc} 
\usepackage[T1]{fontenc}    
\usepackage{hyperref}       
\usepackage{url}            
\usepackage{booktabs}       
\usepackage{amsfonts}       
\usepackage{nicefrac}       
\usepackage{microtype}      
\usepackage{lipsum}		
\usepackage{graphicx}
\graphicspath{ {./fig/} }
\usepackage[square,numbers]{natbib}
\usepackage{multirow}
\usepackage{doi}
\usepackage{nomencl}
\usepackage[ruled]{algorithm2e}
\usepackage{float}

\title{Technology Fitness Landscape for Design Innovation: \\ A Deep Neural Embedding Approach Based on Patent Data}

\date{A Preprint, Version: Oct 20, 2022} 					

\author{ 
{
\hspace{1mm}Shuo Jiang}\thanks{Comments are welcome: \texttt{shuojiangcn@gmail.com}} \\
	Shanghai Jiao Tong University\\
	800 Dongchuan Road, Shanghai, China, 200240\\
	\texttt{shuojiangcn@gmail.com} \\
	\And
	{\hspace{1mm}Jianxi Luo} \\
	Singapore University of Technology and Design\\
	8 Somapah Road, Singapore, 487372\\
	\texttt{luo@sutd.edu.sg} \\
}

\renewcommand{\headeright}{Technology Fitness Landscape for Design Innovation}
\renewcommand{\undertitle}{}
\renewcommand{\shorttitle}{}

\begin{document}
\maketitle

\begin{abstract}
	Technology is essential to innovation and economic prosperity. Understanding technological changes can guide innovators to find new directions of design innovation and thus make breakthroughs. In this work, we construct a technology fitness landscape via deep neural embeddings of patent data. The landscape consists of 1,757 technology domains and their respective improvement rates. In the landscape, we found a high hill related to information and communication technologies (ICT) and a vast low plain of the remaining domains. The landscape presents a bird’s eye view of the structure of the total technology space, providing a new way for innovators to interpret technology evolution with a biological analogy, and a biologically-inspired inference to the next innovation.
\end{abstract}


\section{Introduction}
\label{sec1}

In the past decade, artificial intelligence, cloud computing, quantum computing, and 5G communication technologies undergo rapid advances. Meanwhile, tremendous innovations also emerge and gain momentum in traditional technological domains, such as autonomous vehicles \cite{claybrook2018autonomous}, drug discovery \cite{vamathevan2019applications}, and protein structure prediction \cite{jumper2021highly}. Such contemporary innovation phenomena call for new theories and frameworks to explain them, understand the driving forces, and inform future innovation.

Many contemporary design innovations share one characteristic in common: they are based on the synthesis and fusion of different technological domains, which used to be unrelated and separately developed, e.g., artificial intelligence and automobile. The rise of such innovations has ambiguated the boundaries of technological domains and industries. For instance, should an autonomous vehicle \cite{claybrook2018autonomous} be classified as an automotive or an artificial intelligence product? Should DNA data storage \cite{church2012next} be defined as biological or information technology? Should ancient DNA analysis \cite{Hofreiter2001} be classified as anthropology or molecular biology technology? More holistic assessment of technological domains in one integrated space is demanded to help interpret their evolution and find directions for design innovation.

In this study, we apply deep learning-based neural embedding techniques on multimodal patent data to train a unified embedding vector space for 1,757 technology domains, which covers more than 97\% of the whole US patent system. The training encodes both the internal semantic features of patent texts in individual domains and external interdependent information of patent citations across different domains. We further map the estimated technological improvement rates for all domains in the space to construct a technology fitness landscape. The shape of the technology fitness landscape allows for a holistic understanding of the structure and evolution dynamics of the total technology space and a context-aware understanding of the evolution prospects of individual technological domains. This was not previously possible as the technology domains were then analyzed discretely or associated in a single-or low-dimensional space (e.g., citation information only).

Our technology fitness landscape creation draws analogical inspiration from Kauffman’s NK biological fitness landscape for assessing the evolution of genotypes \cite{kauffman1993origins}. It inspired us to link the domain structure of the entire technology space to the genetic structure of the organism, and the adaptive evolution process of technology to the mutation of the genotypes. The NK model has also been successfully used to study the evolution of firms \cite{levinthal1997adaptation} and innovation networks \cite{ganco2017nk}. In our study, a specific technological domain was found to be analogous to a genotype. The high-dimensional embedding vector of a technological domain is analogous to a DNA sequence. The improvement rate of a technology domain is analogous to the fitness (or replication rate) of a genotype in an organism. Overall, the technology fitness landscape is analogous to the genome fitness landscape.

Within the technology fitness landscape, we find that the fastest domains of today’s technological world gather closely in one region of the entire space. That is, the technology fitness landscape shows a single high hill, which is related to information and communication technologies (ICT). The hill is steep. The improvement rates drop rapidly from the global peak to the low plain. Most regions of the total technology embedding space are in a vast low-flat plain in terms of improvement rates. The entire landscape map can inform innovators within specific domains about their embeddedness in the total space and guide them to mutate their designs or technologies for innovation toward the direction of interest.

This study contributes to the growing studies of design innovation \cite{Li2021,Fiorineschi2021}, computer-aided innovation (CAI) \cite{Luo2021,han2018computational,Chen2019}, and data-driven innovation (DDI) \cite{Luo2022,Jiang2021datadriven,Jiang2022}. The remainder of this paper is organized as follows. Section 2 describes the method and data that we used to create the technology fitness landscape. Section 3 presents the results with our analysis and discussion. Section 4 discusses the findings and applications. Finally, Section 5 concludes the paper.

\section{The Creation of the Technology Fitness Landscape}
\label{sec:sec2}

\subsection{Method}

To construct the ‘map of technologies’, previous studies leveraged the information in patent data to capture the relationships among different technological domains in the form of networks \cite{Luo2019,Acemoglu2016,fleming2004science}. However, these existing maps only analyze the citation information of patents to capture the interdependencies or interactions among technological domains while ignoring the intrinsic features of technologies within them. In this research, we leverage deep learning techniques to create a compact, dense, and continuous vector space in high dimensions to represent all technologies based on multimodal information (both texts and citations of patents). Both the internal (semantic) and external (connective) information of technological domains are utilized for neural embedding model training, aiming to capture both the explicit and implicit features of technological domains. The trained technology embedding space is more holistic and informative than the previously created maps because such a space is created from a distributed high-dimensional representation of items. The representation learning process based on the graph neural network is more flexible and robust than feature engineering \cite{levy2015improving}.

Specifically, our proposed neural embedding method builds on the GraphSAGE model, a framework for inductive representation learning on large graphs \cite{hamilton2017inductive}. GraphSAGE generates vector representations for nodes and is especially useful for graphs with rich node attribute information. In the GraphSAGE model, node embeddings are learned by solving a simple classification task: given a large set of ‘positive’ (target, context) node pairs generated from random walks performed on the graph (i.e., node pairs that co-occur within a certain context window in random walks), and an equally large set of ‘negative’ node pairs that are randomly selected from the graph according to a certain distribution, the model can be trained as a binary classifier to predict whether arbitrary node pairs are likely to co-occur in a random walk performed on the graph. Through learning this simple binary node-pair-classification task, the model automatically learns an inductive mapping from attributes of nodes and their neighbors to node embeddings in a high-dimensional vector space, which preserves structural and feature similarities of the nodes. It is noteworthy that this neural embedding method is an unsupervised learning algorithm that does not require any node class label data for training.

Algorithm 1 describes the GraphSAGE neural embedding method. The spirit behind this algorithm is that as the nodes learn features from their local neighbors, they would incrementally obtain an increasing amount of information from further reaches of the whole graph.

\begin{algorithm}[H]
    \caption{GraphSAGE neural embedding algorithm from \cite{hamilton2017inductive}}
    \label{alg:algorithm1}
    \SetKwInOut{Input}{Input}
    \SetKwInOut{Output}{Output}
    \Input{Graph $\mathcal{G}(\mathcal{V},\mathcal{E})$;
    input features$\{x_v,\forall v \in \mathcal{V}\}$;
    depth $K$;
    weight matrices $\mathcal{\textbf{W}}^k, \forall k \in \{1,...,K\}$;
    non-linearity $\sigma$;
    differentiable aggregator functions $\textsc{AGGREGATE}_k, \forall k \in \{1,...,K\}$;\\
    neighborhood function $\mathcal{N}:v\rightarrow2^\mathcal{V}$
    }
    \Output{Vector representations $\textbf{z}_v$ for all $v\in \mathcal{V}$}
    
    \BlankLine

    $\textbf{h}^0_v \leftarrow \textbf{x}_v, \forall v \in \mathcal{V}$ \\
    \For{$k=1...K$}{
    	  \For{$v \in \mathcal{V}$}{
    	  $\textbf{h}^{k}_{\mathcal{N}(v)} \leftarrow \textsc{aggregate}_k(\{\textbf{h}_u^{k-1}, \forall u \in \mathcal{N}(v)\})$\;
    	  		$\textbf{h}^k_v \leftarrow \sigma\left(\textbf{W}^{k}\cdot\textsc{concat}(\textbf{h}_v^{k-1}, \textbf{h}^{k}_{\mathcal{N}(v)})\right)$
    	  }
    	  $\textbf{h}^{k}_v\leftarrow \textbf{h}^{k}_v/ \|\textbf{h}^{k}_v\|_2, \forall v \in \mathcal{V}$
    	}
     $\textbf{z}_v\leftarrow \textbf{h}^{K}_v, \forall v \in \mathcal{V}$ 

\end{algorithm}

In the experimental setting, we use the domain-level citation network with normalized weights as the input graph $\mathcal{G}$. Formally, an entry in matrix $\textbf{W}_{N \times N}$ from a citing domain $n$ (row) to a cited domain $n'$ (column) is: 

$$ w_{n\rightarrow n'}=\frac{C_{n\rightarrow n'}}{\sum^{N}_{K=1} C_{n\rightarrow k}}$$

where $C_{n\rightarrow n'}$ represents the citations from domain $n$ to domain $n'$. Thus, the normalized weight matrix represents interdomain interactions. 

For intrinsic characteristics of patents, we use semantic vectors of domains as input node features derived from the Doc2Vec model \cite{le2014distributed}. The model is first trained on the text of the title and abstract of all patents in our dataset. Using the trained model, for any given patent document, we can obtain its semantic vector using its textual content. Then, the domain semantic vectors are calculated by averaging all the obtained semantic vectors of patents that belong to the corresponding domains. As for the parameters of the GraphSAGE model, we implemented the settings as suggested in the original GraphSAGE paper: non-linearity $\sigma$ = sigmoid function; aggregator function = Mean; Depth $K$ = 2 with neighborhood sample sizes = 32. Thus, the derived multimodal domain embeddings can effectively encode both the internal semantic and the external connectivity characteristics of patent domains.

\subsection{Data}

The dataset contains 4,988,929 patents with 54,798,218 citations issued by the USPTO from 1976 to 2015, which covers 97.2\% of the entire US patent system during the period. Detailed patent information, including semantic information (titles and abstracts) and connective information (forward and backward citations) of patents, can be downloaded from the PatentsView website (https://www.patentsview.org/).

According to Singh et al.’s prior work \cite{singh2021technological}, all patents had been assigned to a set of 1,757 technology domains via the extended classification overlap method \cite{benson2015technology}. They also statistically estimated the yearly rates of performance improvement for all domains utilizing the centrality measure \cite{triulzi2020estimating}. On this basis, we applied our multimodal neural embedding method to a citation network of 1,757 domains with their semantic features. As a result, we obtained a 32-dimension vector for each of the 1,757 technological domains. We call the unified vector space the technology embedding space. The technology fitness landscape is built on the technology embedding space and technology improvement rates of all domains.

\subsection{Model Training and Validation}

In the training process, we trained our neural network model with the following settings. First, we built a 2-layer GraphSAGE model, and the size of the GraphSAGE layer is set to 32. We stacked the GraphSAGE and prediction layers and defined the binary cross-entropy as the loss function. The entire model is trained end-to-end by minimizing the loss function of choice (e.g., binary cross-entropy between predicted node pair labels and true link labels) using stochastic gradient descent (SGD) updates of the model parameters, with minibatches of ‘training’ links generated on demand and fed into the model. Node embeddings are finally obtained from the encoder part of the trained classifier. Table 1 presents the hyperparameters of the model.

\begin{center}
\begin{table}[!htb]
	\caption{Hyperparameters of the model}
	\centering
	\begin{tabular}{p{6cm}<{\centering}p{4cm}<{\centering}}
		\toprule
		\textbf{Hyperparameters}    & \textbf{Setting values}\\
		
		\midrule

            Number of walks per node & 200 \\
            Walk length & 5 \\
            Batch size & 50 \\
            Training epochs & 10 \\
            Size of GraphSAGE layer & 32 \\
            Learning rate of SGD optimizer & 1e-3\\
		
		\bottomrule
	\end{tabular}
	\label{tab:table1}
\end{table}
\end{center}

To assess the trained model, we used the distribution of cosine similarities to validate the derived embedding space (Figure 1) and show the difference among alternative representation methods, including the Node2Vec embedding, the Doc2Vec embedding, and the citation weight embedding. The experimental settings of Node2Vec and Doc2Vec models are the same as the corresponding parts of the GraphSAGE model, and the dimension values of the representations are all set to 32-d. The citation weight embedding represents each domain as a 1,757-d vector and the value of each dimension means the citation count from that domain to the chosen domain.

\begin{figure}
	\centering
	\includegraphics[width=15cm]{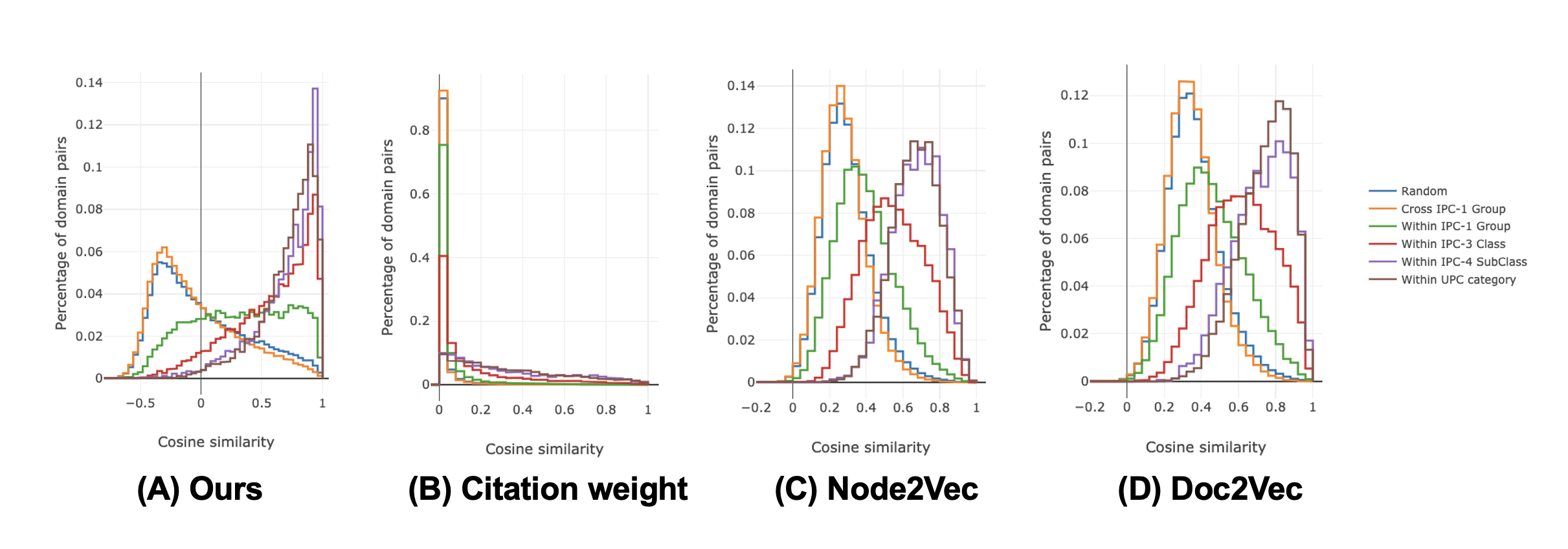}
	\caption{Cosine similarity distribution of domain pairs within different embedding spaces}
	\label{fig:fig1}
\end{figure}

We randomly sampled six groups of 100,000 domain pairs and then filtered out the duplicated ones for further calculation\footnote{IPC denotes the International Patent Classification scheme, and UPC denotes the United States Patent Classification scheme.}: (1) random pairs, (2) cross IPC-1 group pairs, (3) within IPC-1 group pairs, (4) within IPC-3 class pairs, (5) within IPC-4 subclass pairs, (6) within UPC category pairs. The sparse embeddings (citation weight representation) put most pairs at 0 and are not informative as the other three dense embeddings, which better capture both similarities and differences among domains. The Node2Vec and Doc2Vec embeddings show similar patterns: compared to random pairs, both mean values and distributions of the other 5 groups shift more dramatically to the right. Compared to Node2Vec and Doc2Vec, GraphSAGE embedding shows it can better distinguish the fine-grained technology domain pairs.

In addition, we used the t-SNE method to visualize and compare different representations, as shown in Figure 2. Different color denotes one of eight IPC-1 groups. The results also show that the visualization map of the GraphSAGE embedding has a clearer clustering pattern for IPC-1 groups than other embedding methods.

\begin{figure}[H]
	\centering
	\includegraphics[width=15cm]{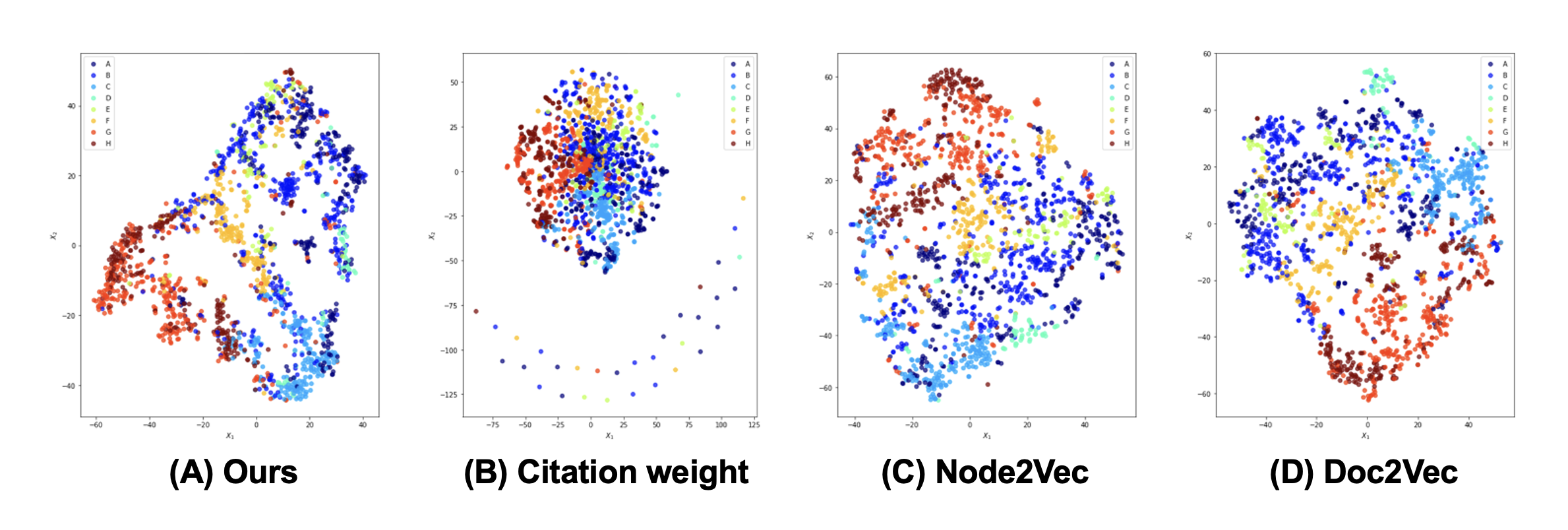}
	\caption{2D Visualization of different embedding spaces via the t-SNE method}
	\label{fig:fig2}
\end{figure}

\section{Results and Analysis}
\label{sec3}

Figure 3A shows a two-dimensional (2D) representation of the technology embedding space using the t-SNE method, where the node color intensity corresponds to the improvement rate of the represented domain. Figure 3B presents the technology embedding space with domain improvement rates as a contour map. The 3D-meshgrid function of MATLAB is used to fit the contour and landscape maps. Both scatter and contour maps reveal a single cluster of the fastest domains, or, the fastest domains tend to gather in a small cohesive region of the total technology embedding space. In other words, our neural embedding space mapping captures the fastest domains within a cluster, including the following domains as the top five:\textbf{719G06F} (dynamic information exchange and support systems integrating multiple channels), \textbf{709G06F} (network management specifically client-server applications), \textbf{709G06Q} (network messaging system including advertisement), \textbf{709H04L} (network address and access management), and \textbf{726H04L} (securing enterprise networks by system architecture). The codes of domains are UPC-IPC pairs \cite{singh2021technological}.

We further developed a web-based interactive visualization for public users to explore the space available at: https://ddi.sutd.edu.sg/technology-fitness-landscape.

\begin{figure}[H]
	\centering
	\includegraphics[width=15cm]{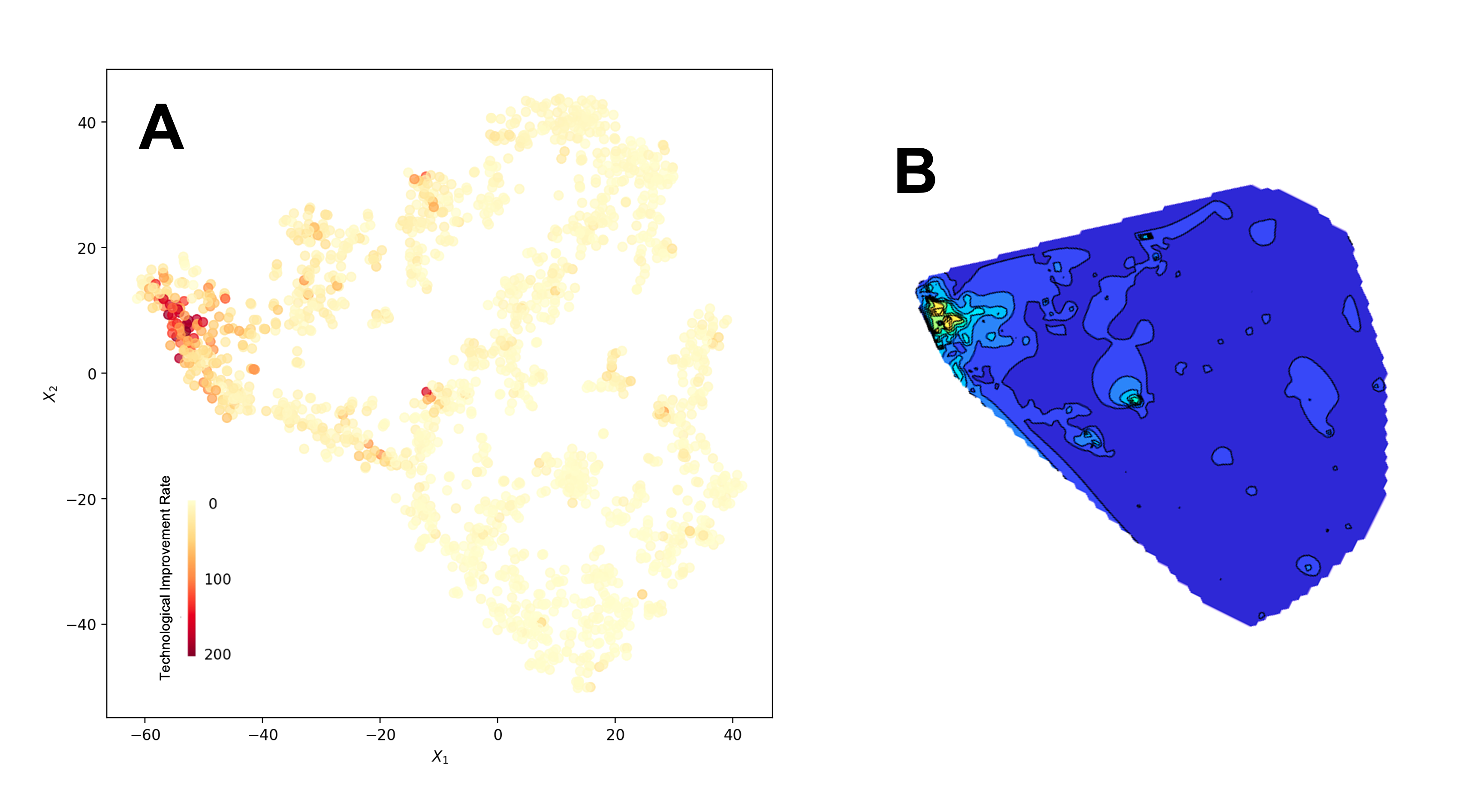}
	\caption{Technology embedding space with improvement rates of all domains. A) Each dot denotes a domain, and its color denotes the domain’s performance improvement rate. The darker dots represent faster domains. B) The lighter areas represent faster domains.}
	\label{fig:fig3}
\end{figure}

Figure 4 presents the technology fitness landscape map, in which the heights of the domains correspond to their respective improvement rates. The landscape is not very rugged and has a small number of hills. The landscape is characterized by a conspicuous highest hill (the global peak), together with a vast low plain occupying most of the total technology space. In Figure 4, we annotate several fastest domains in each hill according to their descriptions. We find that the highest hill is highly related to the information, electronics, and electrical technology domains\footnote{We also performed a Non-negative Matrix Factorization topic modeling on the domains of the global peak to examine its theme. Results show that all of the identified topics are about information, electronics, and electrical technologies, as shown in Figure 7.}. The slope from the peak of the highest hill to the low-flat plain is steep.

\begin{figure}[H]
	\centering
	\includegraphics[width=15cm]{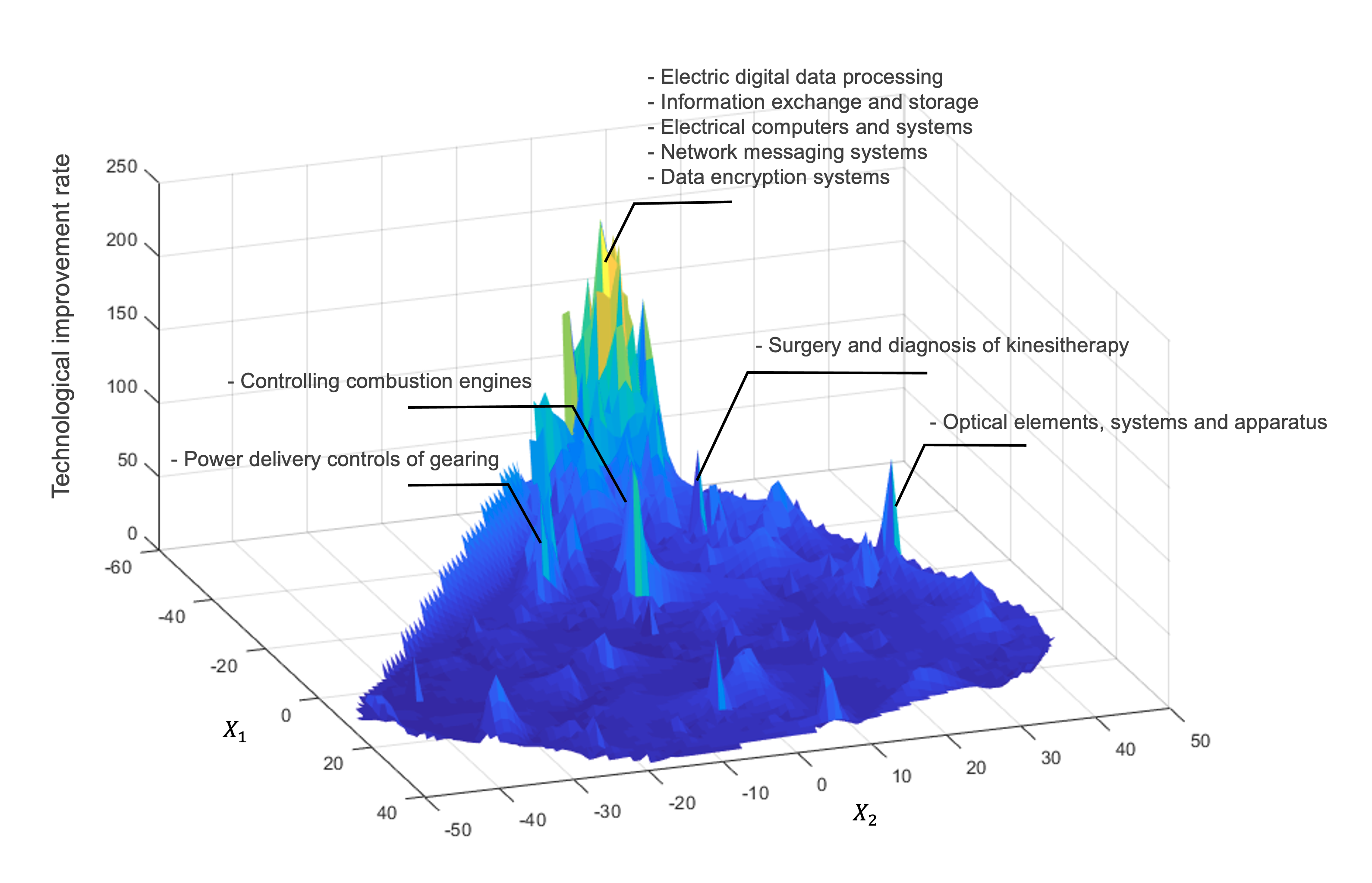}
	\caption{Technology fitness landscape. The location of each domain is aligned to the 2D embedding map (Figure 3A), and the color represents the rate. The heights correspond to the improvement rates of different domains.}
	\label{fig:fig4}
\end{figure}

Figure 5 presents the average improvement rates of each 10-quantiles group of domains by their Euclidean distance to the centroid of the N (=1, 5, or 10) domains surrounding the global peak. The results confirm that, in the near field surrounding the global peak, the improvement rates decline rapidly from the hill peak to the low plain of many slow domains.

\begin{figure}[H]
	\centering
	\includegraphics[width=15cm]{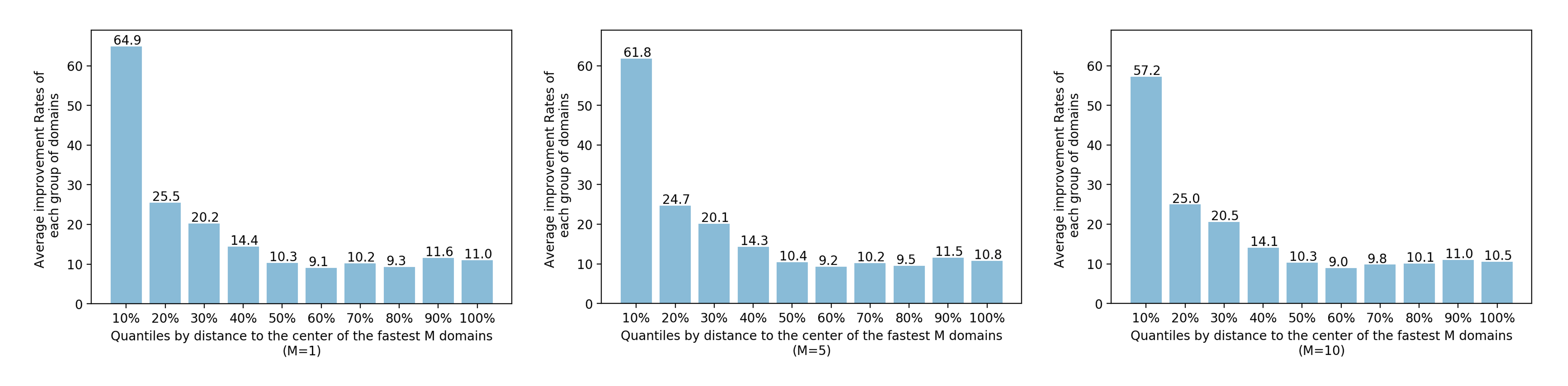}
	\caption{Distance to the global peak (the center of the fastest M domains, M=1, 5, 10) and improvement rates of each group of domains.}
	\label{fig:fig5}
\end{figure}

To demonstrate different domains’ varying distances to the global peak, we mapped all technological domains into one of the 37 NBER subcategories \cite{hall2001nber}. These 37 subcategories belong to six categories: (1) computers and communications, (2) electrical and electronic, (3) mechanical, (4) drugs and medical, (5) chemical, and (6) other. Figure 6 presents the constitution of each category regarding their distances to the global peak (the centroid of the 10 fastest domains). Each item in the matrix represents the number of domains belonging to the corresponding subcategories. The matrix was normalized by column (the sum of each column equals 1). This reveals a clear pattern of technological theme shifts from the global peak to the low plain.

\begin{figure}[H]
	\centering
	\includegraphics[width=15cm]{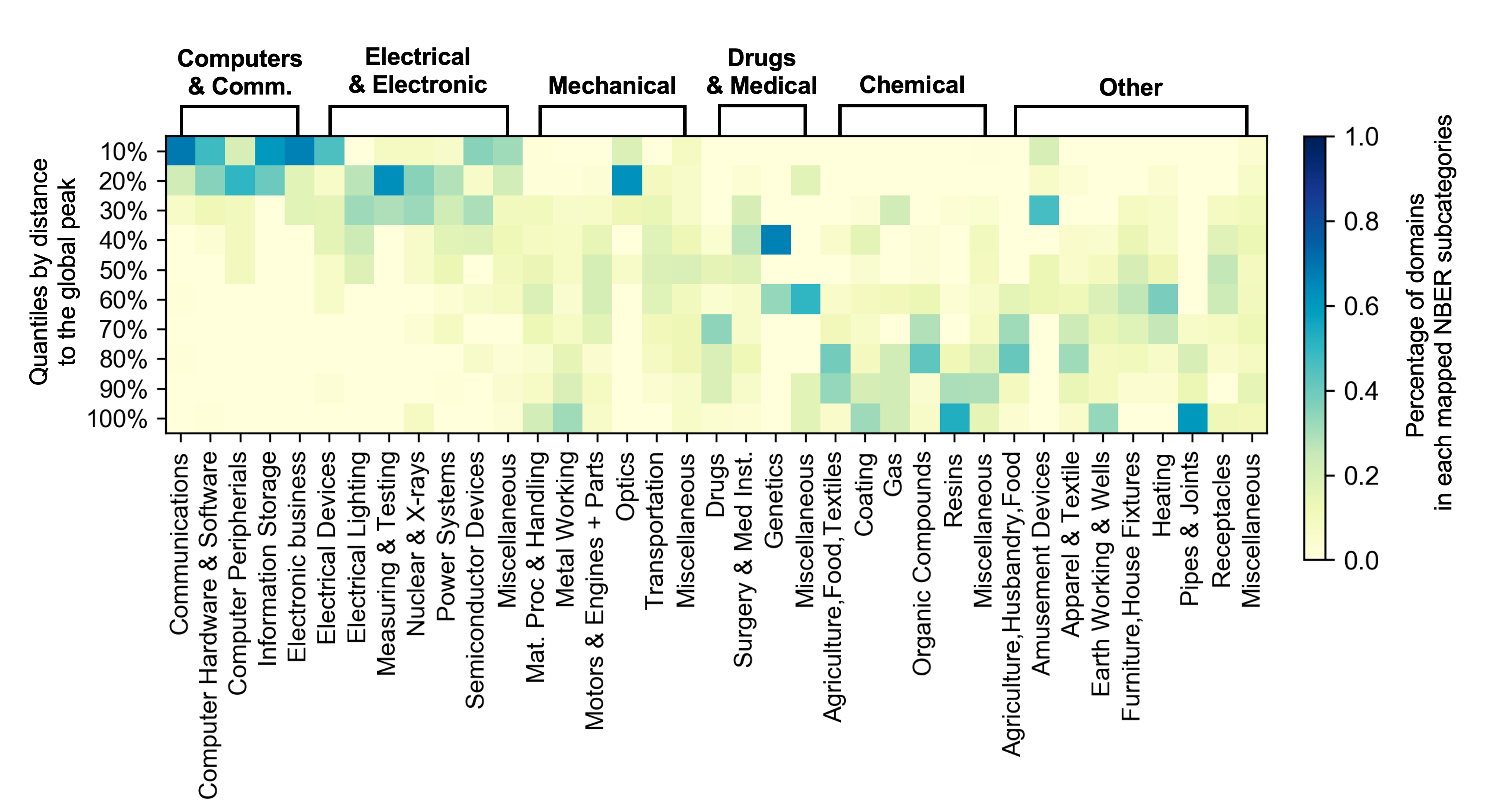}
	\caption{The shift of technological themes of domains in NBER subcategories regarding their distances to the global peak.}
	\label{fig:fig6}
\end{figure}

\section{Discussion}
\label{sec:sec4}

\subsection{Discussion on findings and implications for innovation}

The creation of the technology fitness landscape was inspired by Kauffman’s NK genome fitness landscape for assessing the evolution of genotypes. The genome fitness landscape comprises genotypes similar or dissimilar to each other to different degrees, and the height of an area corresponds to the fitness (or replication rate) of a particular genotype. In the genome fitness landscape, each genotype is composed of several nucleotides (DNA sequences, which can be viewed as 4-d vectors), and each position in a DNA sequence can be occupied by four alternative bases: adenine (A), thymine (T), guanine (G), and cytosine (C). By analogy, the technology domains in the total technology fitness landscape are similar to the genotypes. The domain features can be viewed as the nucleotides in the DNA sequence of the technological domain and are now characterized by our 32-dimensional vectors trained on multimodal data. 

Therefore, technological changes or innovations in a domain are analogous to the mutations of a genotype. In the biological evolution process, mutations in a DNA sequence, which substitute, insert, or delete a single nucleotide, will lead to movements in the genotype landscape and the fitness-increasing mutations can lead to movements to higher areas of the genome fitness landscape. By analogy, the mutations of domain features lead to movements in the technology landscape, and innovations that are those performance-improving mutations can lead to movements from the lower to higher domains in our technology fitness landscape.

Prior studies have pointed out that technological improvement or novelty arises from the recombination or synthesis of existing technologies \cite{he2017novelty,he2019mining}, which, in our cases, can be viewed as such mutations of the technological genotype. Following the analogy framework, the latest innovations in autonomous vehicles have changed the genotype of automobiles and increased the values of automobiles by fusing artificial intelligence to assist or automate driving and battery-powered electric powertrain to replace combustion engines. Similarly, recent progress on the structure prediction component of the ‘protein folding problem’ achieved by DeepMind also presents the power of incorporating deep learning techniques (AlphaFold) into traditional biological domains \cite{jumper2021highly}. In the past, it would take biologists six months to predict a protein structure, while now it takes only a couple of minutes using AI. Speaking in biological evolution terms, these domains’ genotypes have been mutated with increased fitness in the total technological space. The new genotype is positioned closer to the global peak in the technology fitness landscape. 

Another interesting example of such disruptive technology is DNA data storage, which uses synthetic DNA as a medium to store massive quantities of digital information at very high density in the long term \cite{church2012next}. Theoretically, a coffee mug full of synthetic DNA could store the data of the entire world \cite{erlich2017dna}. In our technology fitness landscape, the emergence of DNA data storage has also changed the genotype of the traditional information storage domains by involving advanced synthetic biology technologies and DNA sequencing, bringing a breakthrough in their performance. Such a biologically inspired analysis suggests the need to continually update the classification of technologies, embrace new domain definitions, and redefine the boundaries of technologies.

For the innovators in presently slow-pacing domains, they may use our fitness landscape map as a guide to mutate their technologies (e.g., innovation) for targeted movements toward the high hill. For the innovators in the fast-pacing domains, they may also use our map to identify slow-paced domains as targets to empower by applying their fast-improving technologies over these. Both ways create value despite different starting points. The success of such mutations might be conditioned by the innovators’ starting positions and neighboring domains in the total space. Making long jumps require learning distant technologies and thus are difficult, while they can often increase the possibility of making huge breakthroughs \cite{Luo2021}.

\subsection{Limitations}

This study has several limitations. First, the visualization and analysis in this study rely on the results and domain classifications, and estimated improvement rates. Researchers focusing on the different granularities of technological domains can leverage our method to retrain the model and generate a new landscape based on their dataset. Second, although our neural embedding method learns both intrinsic and connective features from domains and can be more holistic than unimodal spaces, other types of knowledge of domains exist, such as affiliation information and visual images. A desirable multimodal space should consider an even broader range of features.

\section{Conclusion}
\label{sec:sec5}

In this work, we leveraged neural embedding techniques to build a technology fitness landscape based on US patent data to boost design innovation. The technology fitness landscape shows that the fastest domains gather closely in the total technology embedding space, while most regions of the space constitute a low plain in terms of improvement rates. A global peak was identified in the landscape with the themes of information and communication technologies. Such a technology fitness landscape provides us with a birds-eye view of the evolution prospects of individual technological domains. The landscape also allows for a biological analogy to understand technological innovation and evolution and to guide innovators in the search for future opportunities and directions of design innovation.

\section*{Disclosure statement}

No potential conflict of interest was reported by the author(s).

\section*{Acknowledgements}
This work was supported by the SUTD-MIT International Design Center, SUTD Data-Driven Innovation Laboratory (DDI, https://ddi.sutd.edu.sg/), and Shanghai Jiao Tong University under the grant of the National Natural Science Foundation of China (52035007, 51975360), Special Program for Innovation Method of the Ministry of Science and Technology, China (2018IM020100), and National Social Science Foundation of China (17ZDA020).

\bibliographystyle{unsrtnat}
\bibliography{references}

\section*{Appendix}
\subsection*{Topic modeling for themes of the domains of the global peak}

We apply a topic modeling algorithm named Non-negative Matrix Factorization (NMF) \cite{lee1999learning} to further analyze the essential topics of the domains of the global peak. The NMF method learns topics by directly decomposing the term-document matrix, which is a bag-of-word matrix representation of a text corpus, into two low-rank factor matrices: a topic-document matrix and a term-topic matrix. Specifically, we fed the abstracts and titles of all patents belonging to the global peak as a corpus into the NMF model to learn topics. We set the number of topics as 10, corresponding to the number of domains. Figure 7 reports the topic summary for the global peak, each represented as a bar plot using the top 10 words based on their associated weightage contribution to the topics. As we can clearly see, all of the identified topics are about information, electronics, and electrical technologies. For example, Topic \#1 shows terms associated with information processing and related apparatus, as evident with the terms ‘information’, ‘processing’, and ‘apparatus’.

\begin{figure}[H]
	\centering
	\includegraphics[width=15cm]{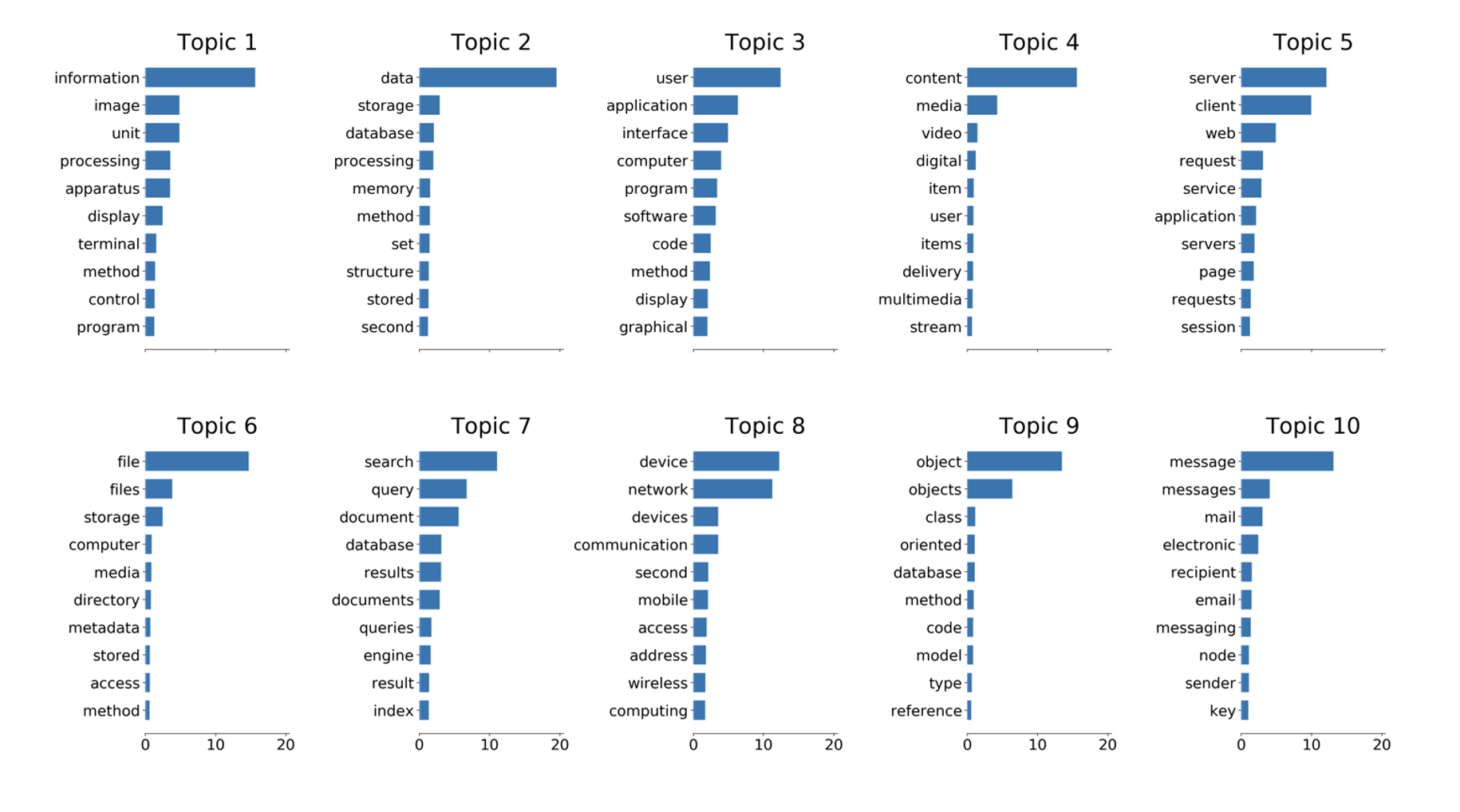}
	\caption{Topic summary for the global peak (the fastest 10 domains). In each topic, ten key-terms are presented with weights.}
	\label{fig:fig7}
\end{figure}

\end{document}